\documentclass[12pt,oneside]{article} 

\usepackage{booktabs}

\usepackage[comma]{natbib}
\bibpunct[, ]{(}{)}{;}{a}{}{,}

\usepackage{authblk}

\usepackage[colorlinks, linkcolor = black, citecolor = black, filecolor = black, urlcolor = black]{hyperref}
\usepackage{graphicx}
\usepackage{subfig}
\usepackage{amsmath,amsthm}
\usepackage{amssymb}
\usepackage{titlesec}
\usepackage{tikz}
\usepackage[text={6.3in,9.5in},a4paper]{geometry}
\usepackage{setspace}
 \usepackage{color}
 \usepackage{colortbl}

\usepackage{fancyhdr}
\usepackage{wasysym}

\usepackage[labelfont={normalsize,bf},font={normalsize,singlespacing},labelsep=period,justification=centering]{caption}

\fancypagestyle{plain}{%
\fancyhead{} 
}
\usepackage{color}

\DeclareMathOperator*{\argmin}{argmin}

\definecolor{blue}{gray}{0}

\newcommand{\customfootnotetext}[2]{{
  \renewcommand{\thefootnote}{#1}
  \footnotetext[0]{#2}}}

\begin{document}

\titleformat{\section}
  {\normalfont\rmfamily\bfseries\color{black}}
  {\thesection}{1em}{}

\titleformat{\subsection}
  {\normalfont\rmfamily\bfseries\color{black}}
  {\thesubsection}{1em}{}


\begin{titlepage}

\title{\Large A Falsificationist Account of Artificial Neural Networks}

\author{\emph{Oliver Buchholz}\textsuperscript{*}\textsuperscript{1}~~and \emph{Eric Raidl}\textsuperscript{*}\textsuperscript{12}\mbox{     }}

\customfootnotetext{*}{University of T\"ubingen, Cluster of Excellence ``Machine Learning: New Perspectives for Science'', T\"ubingen, Germany. E-mail: \href{mailto:oliver.buchholz@uni-tuebingen.de}{\texttt{oliver.buchholz@uni-tuebingen.de}; \href{mailto:eric.raidl@uni-tuebingen.de}{\texttt{eric.raidl@uni-tuebingen.de}}. We would like to thank two anonymous referees as well as the participants of the CEPE/IACAP joint conference 2021 for helpful comments on the initial idea. We would also like to thank Sara Blanco, Karoline Reinhardt and Thomas Grote for helpful feedback on a first draft, Wolfgang Spohn for very valuable comments on an earlier version, and two anonymous referees whose comments helped us improve the article significantly.}}
\customfootnotetext{1}{Funded by the Baden-W\"urttemberg Foundation (program ``Verantwortliche K\"unstliche Intelligenz'') as part of the project AITE (Artificial Intelligence, Trustworthiness and Explainability).}
\customfootnotetext{2}{Funded by the Deutsche Forschungsgemeinschaft (EXC number 2064/1, project number 390727645).}

\maketitle

\thispagestyle{empty}

\begin{abstract}
{\footnotesize
\noindent Machine learning operates at the intersection of statistics and computer science. This raises
the question as to its underlying methodology. While much emphasis has been put on the close link between the process of learning from data and induction, the falsificationist component of machine learning has received minor attention. In this paper, we argue that the idea of \emph{falsification} is central to the methodology of machine learning. It is commonly thought that machine learning algorithms infer general prediction rules from past observations. This is akin to a statistical procedure by which estimates are obtained from a sample of data. But machine learning algorithms can also be described as choosing one prediction rule from an entire class of functions. In particular, the algorithm that determines the weights of an artificial neural network operates by empirical risk minimization and rejects prediction rules that lack empirical adequacy. It also exhibits a behavior of implicit regularization that pushes hypothesis choice toward simpler prediction rules. We argue that taking both aspects together gives rise to a falsificationist account of artificial neural networks. 
}

\end{abstract}

\end{titlepage}

\newpage
\pagestyle{fancy}
\setcounter{page}{2}
\setcounter{footnote}{2} 

\section{Introduction}
\label{sec:introduction}

Machine learning (ML) methods are astonishingly successful in predictive tasks ranging from board games (\citeauthor{Silver.2016} \citeyear{Silver.2016}, \citeyear{Silver.2017}) to medical diagnostics \citep{Esteva.2017} and applications in structural biology \citep{Jumper.2021}.  However there is no definite explanation for this predictive success. An answer to that question ultimately needs to refer to the methodology by which ML algorithms learn. But even this is unclear and not settled, for ML operates at the intersection of statistics and computer science (\citeauthor{Jordan.2015} \citeyear{Jordan.2015}, \citeauthor{Wheeler.2016} \citeyear{Wheeler.2016}).

Classical statistics is about drawing inferences from finitely many observations. Consequently, the focus of classical statistics has been on inductively estimating parameters \citep{Fisher.1925} and providing theoretical guarantees for their reliability  \citep{Breiman.2001}. Briefly put, the more data, the more reliable the statistical inferences.\footnote{
The classical results are the Law of Large Numbers and the Central Limit Theorem.}

\textcolor{blue}{Classical} computer science deals with the design of algorithms that solve particular problems by following a given set of rules (\citeauthor{Uspensky.1993} \citeyear{Uspensky.1993}: ix) in a stepwise procedure \citep{Kolmogorov.1953}.\footnote{For other notions of an algorithm, see \citeauthor{Angius.2021} (\citeyear{Angius.2021}: Sect. 3).} The rule-following notion of an algorithm is closely related to logic. Given a certain number of inputs, the algorithm treats it as premisses, and attempts to find a conclusion by deductive means.

\textcolor{blue}{Despite this apparently sharp initial distinction, both disciplines have been getting increasingly close. The field of ML is a paradigmatic example for this development, being located at a focal point at which progress is due to researchers from both camps.} This opens a number of debated questions. From the perspective of philosophy of science: what is the scientific status of ML? Can it be considered a science?\footnote{See \citet{Gillies.1999} and Williamson (\citeyear{Williamson.2004}, \citeyear{Williamson.2010}).} More precisely, what is its underlying methodology? Is it driven by induction or deduction, or a `combination' of these? 

The standard answer stresses the statistical side of ML and argues that ML algorithms inductively infer general prediction rules from observations 
(\citeauthor{Hastie.2009} \citeyear{Hastie.2009}: 1, \citeauthor{Luxburg.2011} \citeyear{Luxburg.2011}: 651).
This paper takes a different route. \textcolor{blue}{This is not to say that we deny the statistical side of ML. Rather, by combining statistical and algorithmic considerations, we argue that the idea of \emph{falsification} is central to ML.}

The process of falsification roughly proceeds as follows (\citeauthor{Popper.1959} \citeyear{Popper.1959}): reject a given hypothesis in light of disconfirming evidence and supersede it with a more `appropriate' one that is as `simple' as possible. \textcolor{blue}{Thus, although being based on the deductive argument that scientific hypotheses can only ever be falsified, a process of falsification proceeds inductively---and hence combines the methodological approaches of deduction and induction. As a consequence, there are several tensions acting on the underlying methodology of ML. First, there is a tension between statistics which is an inherently probabilistic discipline and falsification which rests on a deductive motivation. Second, there is a tension between statistical induction which aims at verification and falsification which aims at rejection of hypotheses. Third, there is also a tension between the statistical goal of accuracy and the falsificationist goal of getting closer to the truth.}

Existing contributions that investigate the relation between ML and falsification focus on an apparent similarity between the \textcolor{blue}{Vapnik-Chervonenkis (VC)}-dimension\footnote{\textcolor{blue}{\textcolor{blue}{The VC-dimension is named after its inventors \citet{Vapnik.1971}. Intuitively, it measures the complexity of a given function class, that is, whether the class contains rather inflexible or flexible functions. For details and a definition, see Section~\ref{subsec:slt}}.}} and Popper's ideas concerning the simplicity of a hypothesis (\citeauthor{Corfield.2009} \citeyear{Corfield.2009}, \citeauthor{Harman.2007} \citeyear{Harman.2007}, 
\citeauthor{Luxburg.2011} \citeyear{Luxburg.2011}: Sect. 10).
It is argued that a low VC-dimension indicates that the class of functions from which the learning process ultimately selects a function is `simple'. Simpler functions are argued to have higher predictive accuracy (\citeauthor{Luxburg.2011} \citeyear{Luxburg.2011}: 660, \citeauthor{ShalevShwartz.2016} \citeyear{ShalevShwartz.2016}: 36-41). This opens a possible parallel to \citeauthor{Popper.1959} (\citeyear{Popper.1959}: Sect. 43), according to whom simpler hypotheses are `more' universal, have more empirical content and thus are easier to falsify.\footnote{To be precise, \citeauthor{Corfield.2009} (\citeyear{Corfield.2009}) as well as \citeauthor{Harman.2007} (\citeyear{Harman.2007}) stress a general similarity between Popper's theory of falsification and the VC-dimension, but draw a distinction between the latter concept and Popper's concept of degrees of falsifiability (introduced in Section \ref{subsec:initial} below).} \textcolor{blue}{Considerations of simplicity can enter the hypothesis selection process in ML either prior to or during the learning process. The first strategy is implemented by imposing an inductive bias, for example in the form of a restriction of the hypothesis class to keep the VC-dimension relatively low. The second strategy is implemented by optimizing not only for accuracy but for a combination of accuracy and simplicity, as in \emph{structural risk minimization} (SRM) or \emph{explicit regularization}\footnote{\textcolor{blue}{We deliberately speak of `explicit' regularization (usually just called regularization) to distinguish it from implicit regularization.}}, where the VC-dimension or a substitute is used to counterbalance the risk of overfitting.}

However, there are three problems with these attempts. First, \textcolor{blue}{exclusively} focusing on the VC-dimension \textcolor{blue}{or other complexity measures emphasizes the role of an overall function class and its specific structure in learning problems; yet it neglects the details of an ML algorithm's stepwise learning process by which one function is ultimately chosen from that overall class.}
Second, the VC-dimension is a class concept---it measures simplicity of a class of functions---, and not a token concept as Popper's simplicity---the simplicity of one hypothesis.\footnote{\textcolor{blue}{This distinction concerns inductive biases and SRM. Explicit regularization uses a token concept of simplicity. For details see Section~\ref{subsec:slt}.}} Third, we argue that recent ML research (\citeauthor{Belkin.2019} \citeyear{Belkin.2019}, \citeauthor{Zhang.2017} \citeyear{Zhang.2017}) suggests that the VC-dimension is not meaningful for artificial neural networks. Thus, the whole line of argumentation that relies on the VC-dimension to carve out the falsificationist component of ML breaks down for one of the most successful ML methods.

\textcolor{blue}{This paper focuses on artificial neural networks (ANNs), although some aspects similarly apply to other ML methods.} The main contribution of the paper is to show that despite the above worries, ANNs can be interpreted in terms of a falsificationist methodology, through another route. This involves two components: a criterion of empirical accuracy that decides on the rejection of hypotheses and a notion of simplicity that guides hypothesis choice. The first criterion rests on insights from statistical learning theory, where ANNs are described as learning according to \emph{empirical risk minimization} (ERM). Put intuitively, the learning algorithm rejects hypotheses with inferior accuracy or higher risk in favor of hypotheses with higher accuracy. The second criterion rests on recent convergence results showing that the learning algorithm of an ANN exhibits a behavior of \emph{implicit regularization} (\citeauthor{Neyshabur.2015} \citeyear{Neyshabur.2015}, \citeauthor{Poggio.2020} \citeyear{Poggio.2020}: Theorem 4). Put intuitively, the learning algorithm tends to prefer `simpler' hypotheses. In the paper, we compare \textcolor{blue}{these characteristics of the} learning process to the methodological falsification \textcolor{blue}{suggested} by \citeauthor{Lakatos.1968} (\citeyear{Lakatos.1968}, \citeyear{Lakatos.1970b}) \textcolor{blue}{within his methodology of scientific research programmes}. It is on this basis that we argue that the learning process of ANNs \textcolor{blue}{implements elements of} a falsificationist methodology. 

\textcolor{blue}{Similar things might be said about other ML methods. Indeed, other ML algorithms also pursue the goal of balancing accuracy and complexity. For example, support vector machines are usually optimized to achieve both an accurate classification of the data and a sparse set of support vectors. However, this form of explicit regularization implements the paradigm of SRM, where the goal of balancing accuracy and complexity is built into the optimization problem. It needs to be distinguished from implicit regularization, which happens freely.\footnote{For details, see Section~\ref{subsec:slt}.}  Additionally, it is yet not clear whether a behavior akin to implicit regularization can be found in other ML algorithms.}

The paper is structured as follows: Section~\ref{sec:popper} presents Popper's idea on falsificationism and Lakatos' refined methodological falsificationism. Section~\ref{sec:slt_ann} outlines ANNs and statistical learning theory. We discuss why the VC-dimension approach breaks down for ANNs and in which way implicit regularization provides another route to understanding their predictive success. Section~\ref{sec:argument} carves out four parallels 
that highlight the similarities and crucial differences between methodological falsificationism and the learning process of ANNs.


\section{Falsificationism}
\label{sec:popper}

This section outlines relevant aspects of falsificationism. First we present Popper's initial ideas. Then we discuss the main objections which led to substantial refinements in the form of Lakatos' methodological falsificationism.\footnote{\textcolor{blue}{It should be added that Lakatos did not suggest methodological falsificationism as a stand-alone account of falsification. Rather, it served him as a building block for his \emph{methodology of scientific research programmes} that focuses on the dynamic evolution of science. On this view, falsification is taking place in a protective belt of hypotheses that is formed around the ``hard core'' of a research programme, which itself is not subjected to falsification (\citeauthor{Lakatos.1970b} \citeyear{Lakatos.1970b}: 191)}.}

\subsection{Popper's Initial Idea}
\label{subsec:initial}

A central motivation of Popper's philosophy of science is his attempt to solve the \emph{problem of demarcation} (\citeauthor{Popper.1959} \citeyear{Popper.1959}: 11): to identify a criterion that allows to discern science from `non-science'. The criterion he famously proposed is that of \emph{falsifiability}. In its most basic form, it means that a theory or hypothesis is scientific if it can in principle be refuted by empirical evidence. This argument is based on Popper's observation of a logical asymmetry between verifiability and falsifiability. He argues in the spirit of Hume that inductive verifiability is logically inadmissible, since universal statements, representing scientific theories, cannot be derived from singular observational statements.\footnote{The view that scientific theories and hypotheses are to be analyzed as universal \textcolor{blue}{(i.e., allquantified)} statements has been a guiding idea in the philosophy of science. \textcolor{blue}{Note that probabilistic hypotheses are not universal statements. Thus, Popper's idea of falsification does not directly extend to probabilistic hypotheses. For the problem of falsifying the latter, see \citet{Gillies.1971}.}} However, universal statements may be contradicted by singular statements: one can deduce the falsity of a universal statement from the falsity of an instance. For example, while it is impossible to empirically verify a claim such as `all ravens are black', it is easily falsified by the observation of a white raven, that is, by observing a single contradicting instance. If scientific hypotheses can only ever be falsified, the process of scientific enquiry reduces to a process of falsification. This process consists in rejecting a given hypothesis if it lacks empirical adequacy and in superseding it with a more appropriate one. Thus, a falsificationist methodology needs to spell out when a hypothesis should be rejected and, more importantly, how the rejected hypothesis is replaced by a more appropriate one. We will now take a closer look at both steps.

As for the first step, it needs to be spelt out when a \textcolor{blue}{universal} hypothesis can be considered as falsified. For Popper, the falsification of some initial hypothesis requires a \emph{falsifying hypothesis} (\citeauthor{Popper.1959} \citeyear{Popper.1959}: 66). This is a hypothesis \textcolor{blue}{which Popper implicitly assumes to be formulated in the same language, so that it can be a universal, a negation of a universal or, equivalently, an existential hypothesis. The falsifying hypothesis also needs to be inter-subjectively testable. Most importantly, it} should be incompatible with the initial hypothesis under investigation.\footnote{According to Popper, the falsifying hypothesis does not possess the same status as the initial scientific hypothesis or theory. Rather, it is a tool that draws attention to contradicting instances in the process of falsification.} If there are singular observations that are in line with the falsifying hypothesis, the initial hypothesis can be regarded as falsified. Hence, for Popper, a scientific hypothesis is not directly falsified by \textcolor{blue}{a singular} contradicting observation. Rather, falsification is accomplished by observing instances of a corresponding falsifying hypothesis.\footnote{\textcolor{blue}{It is for this reason, it seems, that Popper does not consider probabilistic falsifying hypotheses: although being incompatible with a given universal hypothesis, they cannot be tested by observing singular instances.}}

As an illustration, consider the universal statement `all ravens are black'. A possible alternative hypothesis is `some ravens are not black'. Clearly, this hypothesis \textcolor{blue}{is from the same language---it is an existential statement. It is also inter-subjectively testable and it} contradicts the initial hypothesis\textcolor{blue}{, since it is equivalent to the negation of the universal statement}. Consequently, it fulfills the requirements of a falsifying hypothesis. To falsify the universal statement that all ravens are black, we would need to encounter at least one instance of the falsifying hypothesis, that is, at least one non-black raven.

We should not be misled by the simplicity of the example. Single observations are often insufficient to give up entire theories. The observation of a single non-black raven may be enough evidence to falsify the given hypothesis, because it does not rely on experiments or complicated scientific instruments. Yet in general, a singular observation might be affected by nonreproducible experimental design, malfunctioning instruments, or measurement error. Popper (\citeyear{Popper.1959}: 66) was aware of this and clearly states that ``non-reproducible single occurrences are of no significance to science''. He tries to address the issue by requiring ``a \emph{reproducible effect} which refutes the theory [or hypothesis]'' (ibid.). However, he does not spell out what exactly constitutes such a reproducible effect.

While the first step in the process of falsification, that is, the rejection of a given hypothesis hinges on the relation between the hypothesis and empirical observations, the second step is less obvious: how to come up with a more appropriate hypothesis after rejecting an inappropriate one? To answer this question, Popper (\citeyear{Popper.1959}: 126) introduces \emph{degrees of falsifiability}, thereby stressing the gradual nature of falsifiability. He argues that the degree of falsifiability is higher for simpler hypotheses. To illustrate the argument, Popper uses an example that involves two hypotheses. The first is that some law of nature has the form of a mathematical function of the first degree, that is, of a linear function; the second is that the law has the form of a mathematical function of the second degree, that is, of a parabola. Popper (\citeyear{Popper.1959}: 127) argues that the first hypothesis is more easily falsifiable than the second, because mathematical functions of the second degree are more complex and contain functions of the first degree as special cases. Thus there are `more' functions that are potentially incompatible with a function of the first degree and hence with the first hypothesis than there are functions incompatible with a function of the second degree and hence with the second hypothesis. This makes the falsification of the first hypothesis easier than the falsification of the second hypothesis. As a consequence, Popper (\citeyear{Popper.1959}: 126) equates the degree of falsifiability and the simplicity of a hypothesis. This gives rise to the following guideline for hypothesis choice: a scientist who rejects some hypothesis as empirically inadequate should supersede it by the simplest hypothesis compatible with the evidence.

\subsection{From Dogmatic to Methodological Falsificationism}
\label{subsec:lakatos}

Clearly, Popper's initial idea contains objectionable aspects. \citet{Popper.1963} proposed several refinements. Further progress is due to Lakatos (\citeyear{Lakatos.1968}, \citeyear{Lakatos.1970b}), who also distinguished between `dogmatic falsificationism' referring to Popper's initial idea, and `methodological falsificationism' referring to a refined version.\footnote{To be precise, Lakatos further distinguishes `na\"ive' and `sophisticated methodological falsificationism'. Here, the focus is on the latter concept.}

Lakatos' critique of dogmatic falsificationism involves three main objections: first, dogmatic falsificationism takes an overly na\"ive position with respect to empirical observation. \citeauthor{Lakatos.1970b} (\citeyear{Lakatos.1970b}: 173) argues that in science, the reliability of empirical observation commonly rests on the reliability of the instrument that is employed to make that observation. For instance, the reliability of observations made using a telescope rests on the reliability of the telescope and on the underlying optical theory (\citeauthor{Lakatos.1970b} \citeyear{Lakatos.1970b}: 179). Thus there is no natural separation between purely theoretical and purely observational propositions as required by dogmatic falsificationists.\footnote{This issue is discussed as the \emph{theory-ladenness}  of observation in the philosophy of science.}

Second, Popper's logical argument for falsification rests on the truth of an observational hypothesis from which the falsity of the universal hypothesis under investigation is deduced. This is problematic ``[f]or the truth-value of the `observational' propositions cannot be indubitably decided'' (\citeauthor{Lakatos.1970b} \citeyear{Lakatos.1970b}: 173). Hence, \citeauthor{Lakatos.1970b} (\citeyear{Lakatos.1970b}: 179) argues that dogmatic falsificationists conflate logical disproof of a hypothesis with mere rejection.

Third, \citeauthor{Lakatos.1970b} (\citeyear{Lakatos.1970b}: 174) observes that some scientific theories are simply not falsifiable. This is because theories are frequently intertwined with auxiliary hypotheses that are introduced in scientific practice.\footnote{\citeauthor{Lakatos.1970b} (\citeyear{Lakatos.1970b}: 175) treats \emph{ceteris paribus} clauses in a similar manner.} It is thus a theory in combination with the auxiliary hypotheses that is tested empirically. However, scientists might adjust the auxiliary hypotheses such that contradictory observations are accomodated. Consequently, any attempt to falsify a theory might be repelled by simply adjusting the auxiliary hypotheses (\citeauthor{Lakatos.1970b} \citeyear{Lakatos.1970b}: 182, \citeauthor{Popper.1959} \citeyear{Popper.1959}: 20).\footnote{This idea can be traced back to the work of \citeauthor{Duhem.1954} (\citeyear{Duhem.1954}). \citeauthor{Quine.1951} (\citeyear{Quine.1951}: 40) famously puts forward a similar point arguing that, ``[a]ny statement can be held true come what may, if we make drastic enough adjustments elsewhere in the system.''}

To mitigate the first and second objection, methodological falsificationism takes into account the methodological decisions---hence its name---made by scientists in practice. These decisions can be regarded as conventions on which the relevant scientific community agrees. Consequently, it is no longer assumed that falsification is achieved through purely observational statements. Rather, falsification is achieved through statements that are agreed upon as `observational', for the necessary devices to make the observation and their theoretical underpinnings are treated as unquestioned background knowledge (\citeauthor{Lakatos.1970b} \citeyear{Lakatos.1970b}: 178, \citeauthor{Popper.1963} \citeyear{Popper.1963}: 238).\footnote{Since the term `background knowledge' refers to implicit assumptions and conventions in scientific practice, it should not be regarded as a kind of knowledge in a narrow epistemological sense.} For instance, when observations are made using a telescope, the underlying optical theory is commonly treated as unquestioned background knowledge. This also has consequences for the truth-value of observational statements: methodological falsificationists acknowledge that in general, it ``cannot be proved by facts but [\dots] may be decided by agreement'' (\citeauthor{Lakatos.1970b} \citeyear{Lakatos.1970b}: 176). Thus, although inspired by Popper's logical argument in favor of falsification, methodological falsificationism no longer equates `falsification' with `disproof', which is a matter of logic, but rather with `rejection', which is a collective epistemic attitude of the relevant scientific community.  

To mitigate the third objection concerning auxiliary hypotheses that are commonly employed to save the hypothesis under investigation, methodological falsificationism adds a dynamic perspective to the process of falsification: given a sequence of competing hypotheses, a hypothesis is falsified when it is superseded by a hypothesis with higher empirical content in line with the evidence (\citeauthor{Lakatos.1970b} \citeyear{Lakatos.1970b}: 184). Thus, auxiliary hypotheses are acceptable as long as they offer a content-increasing explanation when combined with the hypothesis under investigation; they are not, when they consist of a mere reinterpretation that tries to overcome the contradiction between hypothesis and evidence in a semantical way.

\citeauthor{Lakatos.1970b} (\citeyear{Lakatos.1970b}: 184) summarizes the preceding amendments as follows: ``There is no falsification before the emergence of a better theory.'' It is in this sense that methodological falsificationism refines and synthesizes the two central steps of Popper's initial idea: that there be a criterion for when to reject a given hypothesis (empirical adequacy) and a normative guideline as to how a rejected hypothesis should be replaced (aim for simpler hypotheses with higher empirical content).


\section{Artificial Neural Networks and Statistical Learning Theory }
\label{sec:slt_ann}

This section provides a short overview of central aspects of ANNs and statistical learning theory: the architecture of ANNs, empirical risk minimization, and the problems of overfitting and underfitting. We close by comparing two approaches to the problem of overfitting: the standard approach based on complexity measures such as the VC-dimension and a new approach based on implicit regularization.

\subsection{Artificial Neural Networks}

ANNs are among the most popular and successful methods in modern ML.\footnote{An entire subfield, deep learning, is concerned with their investigation (\citeauthor{Goodfellow.2016} \citeyear{Goodfellow.2016}, \citeauthor{LeCun.2015} \citeyear{LeCun.2015}).} ANNs are commonly depicted as graphs consisting of nodes, so-called neurons, and edges that transmit the output of one neuron to the input of another neuron. For simplicity, we focus on feedforward networks in which there are no cycles in the graph.\footnote{We follow the notation in \citeauthor{ShalevShwartz.2016} (\citeyear{ShalevShwartz.2016}: ch. 20).}

\begin{figure}
\centering
\def\layersep{2.8cm}

\begin{tikzpicture}[shorten >=1pt,->,draw=black!50, node distance=\layersep]
    \tikzstyle{every pin edge}=[<-,shorten <=1pt]
    \tikzstyle{neuron}=[circle,minimum size=22pt,inner sep=0pt]
    \tikzstyle{input neuron}=[neuron, fill=black!80];
    \tikzstyle{output neuron}=[neuron, fill=black!20];
    \tikzstyle{hidden neuron 1}=[neuron, fill=black!60];
    \tikzstyle{hidden neuron 2}=[neuron, fill=black!40];
    \tikzstyle{annot} = [text width=7em, text centered]

    \foreach \name / \y in {1,...,4}
        \node[input neuron, pin=left: Input \y] (I-\name) at (0,-\y) {};

    \foreach \name / \y in {1,...,5}
        \path[yshift=0.5cm]
            node[hidden neuron 1] (H1-\name) at (\layersep,-\y cm) {};

    \foreach \name / \y in {1,...,5}
        \path[yshift=0.5cm]
            node[hidden neuron 2] (H2-\name) at (2*\layersep,-\y cm) {};

    \node[output neuron,pin={[pin edge={->}]right: Output}, right of=H2-3] (O) {};

    \foreach \source in {1,...,4}
        \foreach \dest in {1,...,5}
            \path (I-\source) edge (H1-\dest);

	\foreach \source in {1,...,5}
        \foreach \dest in {1,...,5}
            \path (H1-\source) edge (H2-\dest);	
            
    \foreach \source in {1,...,5}
        \path (H2-\source) edge (O);

    \node[annot,above of=H1-1, node distance=1.2cm] (hl1) { Hidden layer\\ $V_1$};
    \node[annot,above of=H2-1, node distance=1.2cm] (hl2) { Hidden layer\\ $V_2$};
    \node[annot,left of=hl1] { Input layer\\ $V_0$};
    \node[annot,right of=hl2] { Output layer\\ $V_3$};
\end{tikzpicture}
\caption{Structure of a feedforward ANN with depth $T = 3$.}
\label{fig:ann}
\end{figure}
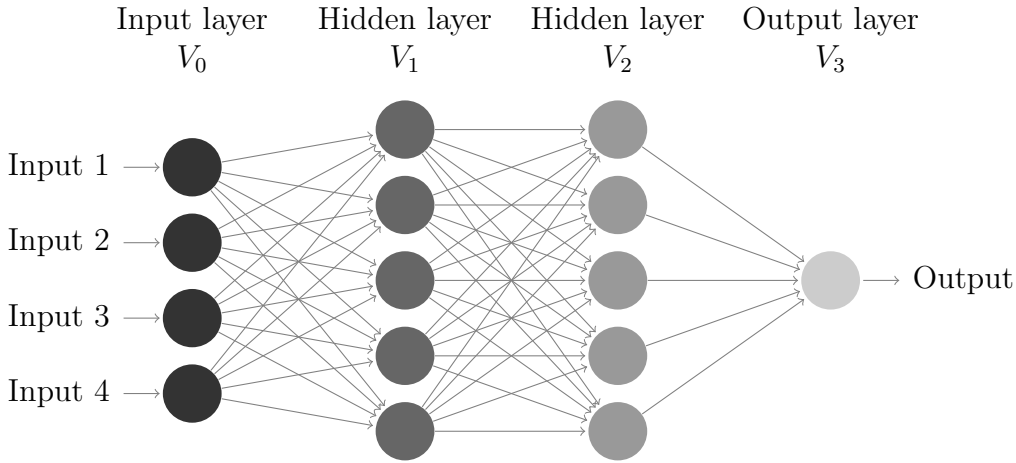

More formally, an ANN can be described as a (directed and acyclic) graph, $G = \langle V,E\rangle$, where $V$ denotes the set of neurons and $E$ denotes the set of edges. Additionally, an ANN is assumed to be structured in layers such that each node from layer $V_{t-1}$ is connected to each node from the next layer $V_t$ by one edge for some $t \in \{1, \ldots, T\}$ (see figure \ref{fig:ann}). Here, $T$ denotes a network's number of layers or \emph{depth}. Input data enters the network at the input layer $V_0$. Then, it is processed through the network as follows: an activation function, $\sigma \colon \mathbb{R} \rightarrow \mathbb{R}$, acts as a filter and transforms the information passed on from one neuron to another.\footnote{Common activation functions are the \emph{rectified linear unit}, $\sigma(a) = \max\{0,a\}$, and the \emph{sigmoid function}, $\sigma(a) = 1/(1 + \exp(-a))$ (\citeauthor{Goodfellow.2016} \citeyear{Goodfellow.2016}: 171, \citeauthor{ShalevShwartz.2016} \citeyear{ShalevShwartz.2016}: 269).} A weight function, $w \colon E \rightarrow Y$, determines, for each edge, the amount of information passed on along that edge. Thus, a neuron's input is the sum of the transformed outputs of all nodes connected to it, weighted by the corresponding weights. Finally, at the output layer $V_T$, the network outputs a label $y$ for each input data $x$.

In practice, researchers usually predefine a graph together with an activation function in form of an \emph{architecture}, $\langle V,E,\sigma \rangle$. The output labels thus depend on the architecture and on the weights, $w$. Since the architecture is fixed, an ANN can be expressed as a function $h_{V,E,\sigma,w} \colon X \rightarrow Y$ that only depends on $w$. Therefore, we abbreviate it by $h_w$.

The goal of `learning' an ANN is to find the best among all possible configurations of weights for a given architecture. This best configuration of weights is then used to predict new labels $y$ for previously unseen input data $x$. As mentioned above, ANNs achieve astonishing predictive success across a wide range of applications. Yet there is no definite explanation for their success. Statistical learning theory provides one possible explanation.

\subsection{Statistical Learning Theory}
\label{subsec:slt}

Statistical learning theory is concerned with the theoretical foundations of ML. It deals with the problem of generalization: How to make accurate predictions for new instances based on empirical observations? In particular, its primary aim consists in providing statistical guarantees for learning algorithms.\footnote{For a concise overview, see \citeauthor{Luxburg.2011} (\citeyear{Luxburg.2011}). For detailed developments see \citeauthor{ShalevShwartz.2016} (\citeyear{ShalevShwartz.2016}) or \citeauthor{Vapnik.1998} (\citeyear{Vapnik.1998}, \citeyear{Vapnik.2000}).} In the following, we focus on supervised learning.

The formal framework of statistical learning theory relies on a rather sparse set of assumptions. The first assumption is static: there is an input space, $X$, and an output space, $Y$. Points in $X$ are possible observations and points in $Y$ are labels usually represented by real numbers (in regression tasks) or categorizations (in classification tasks). Additionally, it is assumed that there is a predetermined class of functions of the type $h \colon X \rightarrow Y$. This class, $\mathcal{H}$, is called the \emph{hypothesis class}. Each function in that class takes a previously unseen observation $x$ and predicts a label $y$. This is why these functions are commonly called \emph{prediction rules}. \textcolor{blue}{A prediction rule $h$ can be read as a collection of universal statements, one for each input point $x$: any individual instantiating the features described by the point $x$ has the property described by the predicted label $h(x)$.} The second assumption is dynamic: the process of learning consists in finding a prediction rule $h$ from $\mathcal{H}$ which is as accurate as possible for the training data. The third assumption is probabilistic: there is a not otherwise specified joint probability distribution $P$ over $X \times Y$. The fourth assumption is statistical: to come up with a prediction rule, an ML algorithm has access to a set of training data. The training data consists of a collection of input-output pairs, $\langle x_1,y_1\rangle, \ldots, \langle x_n,y_n\rangle$.\footnote{This is particular to supervised learning as opposed to unsupervised learning, where labels are not accessible for the algorithm.} It is assumed that the training data is created by some data-generating process, usually by i.i.d.-sampling. The latter means that all input-output pairs are independent from each other and drawn from the same probability distribution $P$. 

To obtain predictions for new data, one final prediction rule needs to be chosen from the hypothesis class. In most cases, ERM or some variant guides this choice. The empirical risk of a prediction rule $h$ on the training data is computed according to

\begin{align}
R_n(h)= \frac{1}{n} \sum_{i=1}^{n} \ell \left(x_i, y_i, h(x_i)\right).
\label{eq:erm}
\end{align}

Here, $\ell(\cdot)$ is a loss function that compares the true label $y_i$ of a training data point to the predicted one, $h(x_i)$.\footnote{Common loss functions are the \emph{0-1-loss}, $\ell \left(x_i, y_i, h(x_i)\right) = 1$ if $h(x_i) \neq y_i$ and $\ell \left(x_i, y_i, h(x_i)\right) = 0$ otherwise, and the \emph{squared loss}, $\ell \left(x_i, y_i, h(x_i)\right) = \left(y_i - h(x_i)\right)^2$ (\citeauthor{Luxburg.2011} \citeyear{Luxburg.2011}: 656).} Since the empirical risk is computed by taking the average across all data points, it measures the average deviation of the predictions from the training data. It is in this sense that the empirical risk can be regarded as a measure of accuracy. Since the goal is to find a maximally accurate prediction rule, the final prediction rule ought to minimize the empirical risk:

\begin{align}
h^* := \argmin_{h \in \mathcal{H}} R_n(h).
\label{eq:final}
\end{align}

This choice is justified by the probabilistic and statistical assumption. The function that yields the most accurate predictions in the training sample will most likely yield the most accurate predictions for previously unseen data. \textcolor{blue}{Of course, this only holds under the assumptions that the training sample contains few misclassifications and that it is representative of unseen data.}\footnote{\textcolor{blue}{The first assumption of correctly labeled data is often implicit while the second assumption is technically captured by identity in distribution. Given these, the above claim is technically} discussed under the term \textcolor{blue}{of} \emph{uniform convergence}, e.g., in \citeauthor{ShalevShwartz.2016} (\citeyear{ShalevShwartz.2016}: Ch. 4).}

Thus, within the framework of statistical learning theory, the learning process in ML can be described as the choice of one final prediction rule from the hypothesis class. The choice is guided by the overall assumption that minimizing the empirical risk over the training data will lead to successful generalization.

\begin{figure}
\centering
\subfloat[Overfitting]{%
      \includegraphics[width=0.45\textwidth]{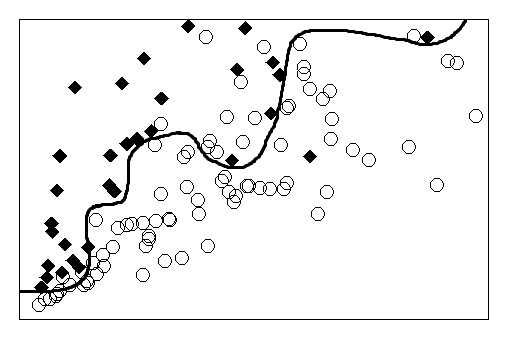}
      \label{subfig:of}
    }
    \subfloat[Underfitting]{%
      \includegraphics[width=0.45\textwidth]{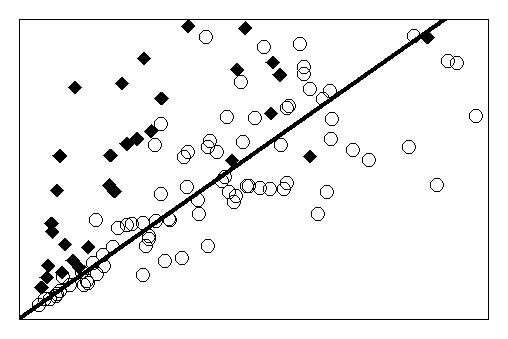}
      \label{subfig:uf}
    }\\
      \subfloat[Balanced situation]{%
      \includegraphics[width=0.45\textwidth]{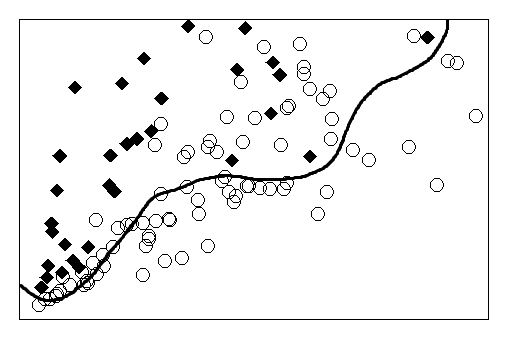}
      \label{subfig:bal}
    }
\caption{The tradeoff between overfitting and underfitting as well as a possible balanced situation in a binary classification task.}
\label{fig:of_uf}
\end{figure}

The ability to generalize \textcolor{blue}{is} closely linked to the two main challenges in ML: underfitting and overfitting. A prediction rule \emph{underfits} when it fits the training data too losely, that is, it achieves poor accuracy on the training data (see \textcolor{blue}{F}igure \ref{subfig:uf}). A prediction rule \emph{overfits} when it fits the training data too closely, that is, it achieves high accuracy on the training data (see \textcolor{blue}{F}igure \ref{subfig:of}). Since the data is sampled randomly, this means that the prediction rule also fits patterns that are specific to the sample at hand, but less relevant for future observations. In both cases, the prediction rule is likely to generalize poorly to new data. 

Underfitting and overfitting do not only concern the dynamic learning process (as driven by ERM), but also the static definition of the hypothesis class, since the final prediction rule is chosen from that class. In fact, whether a prediction rule is more likely to underfit or overfit hinges on the complexity of the underlying hypothesis class. Intuitively, the complexity of the hypothesis class refers to the flexibility of the prediction rules it contains.\footnote{The flexibility of a prediction rule is best illustrated by considering the degree of a mathematical function: a function of the first degree is a straight line and hence rather inflexible. A function of the second degree is a parabola and hence slightly more flexible. Further increasing the degree of the function will also increase the function's flexibility.} A hypothesis class with low complexity contains rather inflexible prediction rules. They may struggle to fit the training data and will be prone to underfitting. A hypothesis class with high complexity contains highly flexible prediction rules. They may even fit random noise in the training data and will be prone to overfitting. Consequently, it is usually necessary to restrict $\mathcal{H}$ in order to balance over- and underfitting.

\textcolor{blue}{ML research has studied two overarching paradigms to implement such restrictions. First, it is possible to incorporate prior knowledge, say, about the structure of the data, when determining the hypothesis class. This restricts the search space for ERM to a specific subset of functions. Such \emph{a priori} restrictions are discussed as \emph{inductive biases} in the literature} (\citeauthor{ShalevShwartz.2016} \citeyear{ShalevShwartz.2016}: 36). For instance, given some indication for a linear relationship between ($d$-dimensional) input and output data, one might assume $\mathcal{H}$ to contain only linear prediction rules. In this case, the hypothesis class would be given by all prediction rules of the following form: $h(x) = x_1 c_1 + \dots + x_d c_d$. The goal of learning the final prediction rule $h^*$ would consist in determining the coefficients $c_1, \dots, c_d$. On the one hand, this restriction would ensure at least an approximate fit between the final prediction rule and the training data. On the other hand, the restriction would prevent the final prediction rule from perfectly fitting the training data. Thus, both underfitting and overfitting would be less likely to occur.

\textcolor{blue}{Second, it is possible to specify a sequence of hypothesis classes $\mathcal{H}_1, \mathcal{H}_2, \mathcal{H}_3, \dots$ with increasing complexity. For instance, this could be achieved by assuming $\mathcal{H}_l$ to be the set of polynomials with degree at most $l$. Then, an algorithm determines $l^* \in \mathbb{N}$ as well as a final prediction rule, $h^* \in \mathcal{H}_{l^*}$, by minimizing the sum of the empirical risk \emph{and} a term that penalizes the complexity of the hypothesis (class). This simultaneous balancing of accuracy and complexity is discussed as SRM in the literature (\citeauthor{ShalevShwartz.2016} \citeyear{ShalevShwartz.2016}: 85, 145).}

\textcolor{blue}{There are different ways to implement SRM in practice.  One way is to use a term $C(l)$ that measures explicitly the complexity of the hypothesis class $\mathcal{H}_l$ to which a prediction rule $h$ belongs.\footnote{\textcolor{blue}{For instance, $C(l)$ could be the VC-dimension of $\mathcal{H}_l$.}} Another way is to use a term $\Omega(h)$ that penalizes the complexity of \emph{that} prediction rule. This is known as regularization (\citeauthor{Luxburg.2011} \citeyear{Luxburg.2011}: 689)---we call it \emph{explicit regularization} to distinguish it from implicit regularization.\footnote{\textcolor{blue}{Many regression techniques (e.g., \emph{ridge regression}) incorporate this strategy. The widely-used (soft margin) \emph{support vector machine} (SVM) also uses such explicit regularization methods (\citeauthor{ShalevShwartz.2016} \citeyear{ShalevShwartz.2016}: 172, 207).}} Whereas the first strategy considers directly the complexity of the hypothesis class, the second considers it through the substitute $\Omega(h)$. However, in both approaches, a term that penalizes complexity is explicitly included in the minimization problem. This distinguishes both approaches from implicit regularization.}

The fact that the tradeoff between over- and underfitting is closely linked to the restrictions imposed on the hypothesis class\textcolor{blue}{, either through the introduction of inductive biases or through a modified optimization problem such as SRM or explicit regularization,} led to the development of a variety of so-called complexity measures in statistical learning theory, the most important being the VC-dimension. These measures are thought to evaluate the complexity of a given hypothesis class, that is, whether it contains rather inflexible functions or functions that are flexible enough to fit a wide variety of data.\footnote{\textcolor{blue}{For instance, the VC-dimension of a hypothesis class is defined ``as the largest number $n$ such that there exists a sample of size $n$ which is shattered by [that class]'' (\citeauthor{Luxburg.2011} \citeyear{Luxburg.2011}: 679). Here, \emph{shattering} means that the hypothesis class can achieve any labeling on the given sample.} For a thorough discussion see \citeauthor{Luxburg.2011} (\citeyear{Luxburg.2011}: Sect. 5.6) as well as \citeauthor{ShalevShwartz.2016} (\citeyear{ShalevShwartz.2016}: Ch. 5).} Thus, it is usually thought that the tradeoff between over- and underfitting reduces to specifying a hypothesis class with the right complexity \textcolor{blue}{or to optimizing for a prediction rule in such a class}. \textcolor{blue}{When this specification is made through inductive biases, it} does not directly affect the dynamic process of choosing the final prediction rule; it rather restricts the \textcolor{blue}{overall} set of prediction rules \textcolor{blue}{to a preferred subset, prior to the learning process,} from which the final choice \textcolor{blue}{is then to be made by learning}. \textcolor{blue}{When this specification is made through SRM or explicit regularization, preferences over the hypothesis class are indirectly expressed by including them in the minimization problem. In both cases, considerations of complexity constrain the learning process and ideally lead to} a final prediction rule at the `sweet spot' between overfitting and underfitting (see \textcolor{blue}{F}igure \ref{subfig:bal}).

\subsection{Peculiarities of Artificial Neural Networks}

At first glance, it seems straightforward to analyze ANNs \textcolor{blue}{just as other ML methods} within the framework of statistical learning theory. Recall, that an architecture, $\langle V,E,\sigma \rangle$, is predefined at the beginning of an ANN's learning process. From the perspective of statistical learning theory, this architecture fixes the hypothesis class of possible prediction rules: it is given by all functions of the form $h_w$ where $w$ is some weight function. More briefly, the hypothesis class is

\begin{align}
\mathcal{H} = \left\lbrace h_w \colon w \textnormal{~is~a~mapping~from~} E \textrm{~to~} Y\right\rbrace. 
\label{eq:hyp}
\end{align}

The hypothesis class is thus determined by the graph and the activation function. Consequently, the functions in the class only differ with respect to their specific configuration of weights. They can be conceived of as different hypotheses about the relation between input and output data.

We have seen that the goal of `learning' an ANN is to find the best configuration of weights for a given architecture. From the perspective of statistical learning theory, `best' means `most accurate', which is why the final configuration of weights is determined via ERM. The actual minimization of the empirical risk is usually achieved using some version of the stochastic gradient-descent (SGD) algorithm. The idea behind the SGD algorithm is to initialize the weights with random values, to update them stepwise and to output the configuration of weights that yields the lowest empirical risk. The process proceeds stepwise as follows: in the first step, the weights in the network are assigned random values, $w_0$. These values determine one specific prediction rule $h_0$ in the hypothesis class. Next, the algorithm computes the empirical risk for this prediction rule, $R_n(h_0)$, by comparing the predicted labels, $h_0(x_i)$, to the true labels $y_i,i = 1, \dots, n$ over the training data. Then, the algorithm runs backward through the network to assess which adjustments to the weights $w_0$ lead to the largest decrease in the empirical risk.\footnote{This process is referred to as `backpropagation' (\citeauthor{Goodfellow.2016} \citeyear{Goodfellow.2016}: Ch. 6.5).} Then the weights are updated accordingly with new values $w_1$. The new weights $w_1$ correspond to a new prediction rule $h_1$. The process is then repeated for $h_1$ and $w_1$. The algorithm stops once the empirical risk cannot be decreased any further. Then, it has found a final configuration of weights $w^*$ that is thought to minimize the empirical risk or, put differently, achieves the highest predictive accuracy.\footnote{While it is possible that SGD converges to a \emph{local} minimum (\citeauthor{Goodfellow.2016} \citeyear{Goodfellow.2016}: 281), there is a high probability for convergence to a \emph{global} minimum in the case of highly complex networks (\citeauthor{Li.2018} \citeyear{Li.2018}, \citeauthor{Poggio.2020} \citeyear{Poggio.2020}: 30044, \citeauthor{Vidal.2017} \citeyear{Vidal.2017}: 2).} This final configuration of weights determines the final prediction rule $h^*$. Therefore, the SGD algorithm is allowed to select \emph{one} prediction rule from the hypothesis class while rejecting all other prediction rules in light of the data. \textcolor{blue}{Until this point, the description of ANNs and their learning process from the perspective of statistical learning theory is very similar to that of other ML methods.}

\textcolor{blue}{I}t is \textcolor{blue}{therefore} straightforward to assume that the tradeoff between over- and underfitting also has a bearing on ANNs. Yet recent results reveal that ANNs possess a remarkable feature: although their architecture is highly complex and they almost perfectly fit the training data, they possess a high ability to generalize to previously unseen data (\citeauthor{Belkin.2019} \citeyear{Belkin.2019}, \citeauthor{Zhang.2017} \citeyear{Zhang.2017}). Put differently, ANNs are generally not susceptible to overfitting. We have seen above that the tradeoff between over- and underfitting is closely linked to the complexity of the hypothesis class as measured, for instance, by the VC-dimension. However, this link seems to dissolve in the case of ANNs. As shown in expression~\eqref{eq:hyp}, their architecture determines their hypothesis class. Thus, a highly complex architecture entails a hypothesis class with high complexity, containing highly flexible prediction rules that can fit a wide variety of data.\footnote{For instance, given only two neurons connected with an edge, the prediction rules are of the form $w\sigma$. If there are several neurons connected to a subsequent neuron, the prediction rules are linear combinations of the activation function. Any further node thus adds to the complexity of prediction rules.} Yet contrary to classical wisdom in statistical learning theory, this high complexity is no longer tied to a susceptibility to overfitting; complex ANNs usually achieve \emph{both} a very close fit to the training data \emph{and} accurate predictions for new observations. Consequently, ML researchers repeatedly argued that conventional complexity measures such as the VC-dimension are not meaningful when analyzing ANNs (\citeauthor{Neyshabur.2017} \citeyear{Neyshabur.2017}, \citeauthor{Zhang.2017} \citeyear{Zhang.2017}). As described above, these measures are thought to evaluate the complexity of a given hypothesis class, thereby offering guidance when trading off overfitting against underfitting. Yet for complex ANNs, these measures would indicate a high risk of overfitting which is not in line with the networks' actual predictive performance.

A general mathematical characterization of the phenomenon remains challenging, which is why many insights about the generalization ability of ANNs still rely on empirical studies \citep{Zhang.2021}. However, there is theoretical progress for several special cases and specific aspects of the problem.\footnote{For instance, \citet{Arora.2019} consider two-layer networks and \citet{Soudry.2018} consider networks with linear activation functions to derive theoretical results.} In particular, recent ML research identified one aspect that might explain the remarkable generalization behavior of ANNs: the SGD algorithm that determines a network's weights has been shown to exhibit a behavior of \textcolor{blue}{\emph{implicit regularization}} (\citeauthor{Neyshabur.2015} \citeyear{Neyshabur.2015}, \citeauthor{Poggio.2020} \citeyear{Poggio.2020}: Theorem 4). This means that the algorithm converges to a final configuration of weights with a small norm.\footnote{A \emph{norm} is a function that takes the elements of a vector as inputs and outputs a non-negative number. It can be interpreted as the `size' of the vector (\citeauthor{Goodfellow.2016} \citeyear{Goodfellow.2016}: 37).} Such a small norm is thought to reflect that the final configuration of weights or the final prediction rule is relatively simple. This is in contrast with the above standard account of statistical learning theory. Whereas, within the standard account, the complexity of prediction rules is restricted \textcolor{blue}{either through imposing inductive biases on the} hypothesis class prior to the learning process \textcolor{blue}{or by following the SRM paradigm,} ANNs \textcolor{blue}{implementing ERM} are different. Here, it is the SGD algorithm that reduces complexity in the process of determining the final configuration of weights\textcolor{blue}{, yet without being told to do so}. Put differently, simplicity of the final prediction rule is usually enforced \emph{ex ante} by the researcher\textcolor{blue}{, either by restricting the hypothesis class (inductive biases) or by including a term that penalizes complexity in the minimization problem (SRM or explicit regularization)}. Yet in the case of ANNs, simplicity is enforced by the SGD algorithm during the learning process. \textcolor{blue}{It is in this sense that regularization is \emph{implicit} in the case of ANNs.}\footnote{\textcolor{blue}{Although explicit regularization is an indirect way of implementing the SRM paradigm, a penalty term is introduced directly in the minimization problem (see Section~\ref{subsec:slt}). It is in this sense that explicit regularization differs from implicit regularization where no such term is introduced.}} In the former case, the simplicity of the final prediction rule is therefore an \emph{ex ante}-simplicity; in the latter case, it is an \emph{ex post}-simplicity.\footnote{\textcolor{blue}{This is not to say that the SRM paradigm is not applicable to ANNs; it is just not the topic of our investigation. See \citeauthor{Goodfellow.2016} (\citeyear{Goodfellow.2016}: Ch. 7) for a treatment of explicit regularization applied to ANNs.}}

In sum, the discussion revealed two central aspects of ANNs: the ERM framework answers the question when the SGD algorithm should reject a configuration of weights in favor of another one; the behavior of implicit regularization sheds light on the algorithm's convergence and hence on the direction it takes when superseding a rejected configuration. Both aspects will become relevant in the sequel.

\section{A Falsificationist Account of Artificial Neural Networks}
\label{sec:argument}

So far, we treated falsificationism and statistical learning theory separately. This section synthesizes the subjects and puts forward the main argument. There are four parallels between the process of finding the final configuration of weights for an ANN and a process of methodological falsification: the first concerns the role of background knowledge; the second concerns the rejection of inappropriate hypotheses; the third concerns the `amount' of disconfirming evidence that is needed to falsify a given hypothesis; the fourth concerns hypothesis choice. \textcolor{blue}{In this section, we discuss these four parallels. However, we also highlight important differences.}

\subsection{The Role of `Background Knowledge'}

As argued by Popper and Lakatos, `background knowledge', that is, implicit assumptions and conventions play an important role in scientific practice. They determine which theories remain unquestioned and which instruments are considered reliable in an enquiry. So while there may be variation in observations or measurements, the instruments employed to observe or measure as well as their theoretical underpinnings remain fixed.

Such implicit assumptions are also of primary importance in the learning process of an ANN.\footnote{\textcolor{blue}{This holds equally for other ML methods.}} Recall that this process starts with the definition of an architecture $\langle V,E,\sigma\rangle$. The architecture is fixed by the choice of a graph and the additional activation function. These choices are based on practical and often implicit knowledge. For instance, there is a certain toolkit of standard activation functions. Overall, a researcher might choose one architecture, since it was already used previously, or by slightly modifying a previously used one. The architecture determines a hypothesis class, $\mathcal{H}$, consisting of all functions $h_w$ associated with the possible weights $w \colon E \rightarrow Y$. While the weights are adjusted during the learning process through ERM, the architecture remains fixed.

This establishes the first parallel: just as particular instruments or theories remain unquestioned and hence fixed in scientific practice, once the network's architecture is chosen, it remains fixed during the learning process of an ANN.

At first blush, this might seem like a difference rather than like a parallel between both processes. After all, scientific instruments and theories are not on a par with the architecture of an ANN. However, the parallel concerns something else. Clearly, the objects that remain fixed---instruments and theories on the one hand, and the architecture on the other hand---are different in kind. Yet what determines these fixed objects is of a similar nature: both scientific practice and the learning process of an ANN are governed by implicit assumptions and conventions that Popper and Lakatos summarize as `background knowledge'. In science, it is background knowledge that justifies the use of certain theories or instruments, which then remain unquestioned during further enquiry. In the context of ANNs it is practical knowledge on which grounds a network's architecture is chosen, that remains unchanged during the learning process.

\subsection{Empirical Risk Minimization as a Process of Falsification}

Recall that usually, the weights of an ANN are determined by the SGD algorithm: starting from a random initialization, the algorithm adjusts the weights iteratively until the empirical risk cannot be decreased any further.

Taking a step back reveals the second parallel: this process is akin to a process of methodological falsification. For each iteration $k$ of the SGD algorithm, there is a clear relation between the new weights, $w_{k+1}$, and those obtained in the previous iteration, $w_k$. The latter correspond to the most accurate prediction rule $h_{k} \in \mathcal{H}$ associated with the lowest empirical risk that could be identified until the current iteration. It serves as a benchmark for subsequent iterations. The new weights, $w_{k+1}$, challenge this benchmark. Thus, the prediction rule $h_{k+1}$ that is determined by the new weights resembles a falsifying hypothesis in a process of falsification: it is incompatible with the previous hypothesis, $h_k$, since obviously $h_{k+1} \neq h_k$, and it is possible to check whether there are instances in the training data that are in line with it. 

In the context of falsificationism, scientific hypotheses are analyzed as universal statements that can be tested by observing singular instances. If there are instances in line with the falsifying hypothesis, the initial hypothesis is regarded as falsified. In the context of ANNs, the process is slightly more complex. \textcolor{blue}{As mentioned in Section~\ref{subsec:slt}, a prediction rule can be read as a collection of universal hypothesis. Thus, it is not the prediction rule itself which is probabilistic, but the way of testing it:} testing \textcolor{blue}{a} falsifying hypothesis amounts to computing its empirical risk, $R_n(h_{k+1})$, over the available training data and comparing it to the risk of the given hypothesis, $R_n(h_k)$. This means that empirical adequacy is summarized in terms of an average accuracy over the training data. If the falsifying hypothesis $h_{k+1}$ produces on average more accurate predictions than the previous hypothesis $h_{k}$, this is regarded as evidence for the falsifying hypothesis. In this case, the old weights $w_k$ are replaced by the new ones, $w_{k+1}$; the hypothesis $h_k$ is falsified and $h_{k+1}$ is the new benchmark. Thus in science and in the learning process of an ANN, a hypothesis is falsified if it lacks empirical adequacy.\footnote{\textcolor{blue}{This insight concerning the underlying methodology of ANNs crucially hinges on the stepwise nature of the SGD algorithm. The existence of distinct iterations allows to investigate the hypothesis involved in each iteration as well as its relation to other hypotheses. Exclusively focusing on the algorithm's output, that is, the final prediction rule, would impede this methodological insight.}}

However, the fact that empirical adequacy is summarized in terms of the empirical risk in the context of ANNs constitutes an important difference. On the one hand, it is only \textcolor{blue}{\emph{one}} possible way of measuring empirical adequacy, on the other hand it makes the meaning of the concept more explicit: a prediction rule is empirically adequate if its empirical risk is low, that is, if it produces on average accurate predictions. \textcolor{blue}{This allows to compare different prediction rules, that is, hypotheses, with respect to their empirical adequacy.}

There is yet another difference: a process of methodological falsification in science is usually driven by the collection of \emph{new} evidence. A given hypothesis is tested and perhaps falsified by observing singular instances that have not been observed before. \textcolor{blue}{According to \citeauthor{Popper.1959} (\citeyear{Popper.1959}: 266), scientists play an active role during this process. They design experiments to collect new data and develop techniques to test a given hypothesis as severely as possible.\footnote{\textcolor{blue}{This aspect is highlighted by \citeauthor{Corfield.2009} (\citeyear{Corfield.2009}: 56).}} This differs from the learning process of an ANN in two ways. First, an ANN is trained on a fixed set of data. There are thus usually no new observations entering the training data during the learning process. Consequently, the empirical risk of hypotheses is evaluated by exploiting the \emph{same} total evidence in each iteration of the SGD algorithm. Second, the fact that the learning process of an ANN is executed by an algorithm reveals that it is an automated, rule-based procedure rather than an active and perhaps even creative endeavor.}

\textcolor{blue}{Whereas previously we saw that ML makes the concept of empirical adequacy more precise, ML can also benefit from ideas of falsificationism. It might be beneficial to extend the predominant learning approach in ML to continuous learning, that is, to a setting in which a learning algorithm is confronted with a constant stream of new data rather than with a fixed training set. Additionally, this continuous learning should not be passive, but actively aiming at confronting the algorithm with new problematic data to enhance the robustness of the final prediction rule.}\footnote{\textcolor{blue}{The field of \emph{online learning} tries to make progress in this direction, studying settings in which training instances become available one by one \citep{ShalevShwartz.2012}.}}

\subsection{\textcolor{blue}{Quantifying a Reproducible Effect}}

Popper highlighted that the falsification of a hypothesis requires a `reproducible effect' that avoids negative impacts of erroneous instances. Although Popper did not elaborate more on this, the idea can be compared to an essential feature in the learning process of an ANN. Here, the empirical adequacy of a hypothesis is evaluated using the empirical risk. The latter is an average over all instances in the training data. That is, it is an average over many observations. It can thus be conceived as measuring the average accuracy of a hypothesis. Since the falsification of a hypothesis is based on the empirical risk, it is thereby based on the consideration of \emph{all} instances in the training data and not only on a singular instance. Considering an average accuracy over several instances reduces the potential for negative impacts of erroneous singular instances. Their effect is assumed to be counterbalanced by the presence of a multitude of correct instances or by the fact that the deviations of singular instances from the truth pull in different directions, such that overall they cancel each other out. This constitutes not only a third parallel, but it also makes Popper's initial idea more precise.

Furthermore, while Popper and Lakatos left the question as to what constitutes a reproducible effect largely unanswered, the `amount' of how much the evidence needs to disconfirm a hypothesis for its falsification can be precisely quantified in terms of the empirical risk within the framework of ERM: given an initial hypothesis $h_k$ and a falsifying hypothesis $h_{k+1}$, falsification of the former is achieved if and only if $R_n(h_{k+1}) < R_n(h_{k})$. \textcolor{blue}{This result emphasizes the importance of the ERM framework in our argumentation. Because ERM evaluates prediction rules exclusively based on their empirical risk, a reproducible effect or the rejection of a prediction rule is directly and only based on available evidence. By contrast, the SRM paradigm focuses on simultaneously optimizing the empirical risk \emph{and} some term measuring the complexity of hypotheses (classes). This makes it hard to ascribe a reproducible effect or the rejection of a prediction rule directly to the evidence. Thus, learning within the SRM paradigm is different from a process of falsification. Indeed, in a process of falsification, evidence is central through the concept of empirical adequacy. Here, it is only after the rejection of one empirically inadequate hypothesis, that considerations of complexity guide the choice of a new hypothesis. While both steps are indistinguishable in SRM, the combination of ERM and implicit regularization allows for this kind of analysis: within the ERM framework, the amount of disconfirming evidence needed for falsification can be precisely quantified and attributed to the instances in the training data when the algorithm evaluates a hypothesis; considerations of complexity are at play after the rejection, when the algorithm picks the next hypothesis.}

Th\textcolor{blue}{e result} also reveals that the framework of ERM is in line with methodological falsificationism for taking the falsification of a hypothesis to be a rejection rather than a disproof. It happens due to insufficient empirical adequacy, not due to strict falsity of the tested hypothesis. \textcolor{blue}{So contrary to Popper's initial account, according to which a process of falsification is directed at getting closer to the truth by rejecting strictly false hypotheses, a process of methodological falsification and the learning process of an ANN seem to pursue the more modest aim of maximizing empirical adequacy. The learning process of an ANN aims at minimizing the empirical risk or, equivalently, at maximizing accuracy.\footnote{\textcolor{blue}{This is evident from the ERM framework and, in particular, from expression \eqref{eq:final}.}} However, accuracy is commonly conceived as a probabilistic replacement for or a gradual notion of truth \citep{Joyce.1998}. It is in this sense that the learning process of an ANN also aims at getting closer to the truth by rejecting inaccurate prediction rules.\footnote{\textcolor{blue}{\citeauthor{Corfield.2009} (\citeyear{Corfield.2009}: 57) also discuss this distinction.}}}

Overall, the process of determining the final configuration of weights within the framework of ERM shares two important aspects with a process of falsification: first, competing hypotheses are assessed according to their empirical adequacy and rejected if they are incompatible with the evidence. The SGD algorithm compares the predictions from the functions in the hypothesis class to the training data and rejects those for which the predictions do not minimize the empirical risk. Second, a hypothesis is only falsified if a reproducible effect in line with an alternative hypothesis could be identified. The empirical risk ensures that this requirement is met, for it is an average over all instances in the training data. In addition to these parallels, there are also important differences: first, empirical adequacy amounts to average predictive accuracy in the context of ANNs. \textcolor{blue}{Second, a process of falsification is usually driven by the collection of new rather than by a fixed set of data. As a consequence, whereas Popper's falsification is driven by an active selection of new data, ANNs learn automatically or passively. Third}, considering an average over many instances to falsify a hypothesis clarifies the meaning of a reproducible effect. 

\subsection{Implicit Regularization as a Preference for Simple Hypotheses}

We have seen above that a process of falsification consists of some criterion to reject a given hypothesis and some guideline for how to replace it with a new one. The criterion for rejection is commonly taken to be empirical adequacy. In ML this is measured by the empirical risk of a prediction rule. The falsificationist guideline for hypothesis choice is simplicity, because it is argued that it is easier to falsify simpler hypotheses. One should therefore replace a falsified hypothesis by the simplest hypothesis compatible with the evidence. We have already mentioned that earlier attempts to carve out the falsificationist component of ML try to connect this guideline for hypothesis choice with the VC-dimension (\citeauthor{Corfield.2009} \citeyear{Corfield.2009}, \citeauthor{Harman.2007} \citeyear{Harman.2007}). Yet we have also shown that this line of argumentation is a dead end in the case of ANNs.

We think nonetheless that there is a parallel to draw. As discussed, recent research on ANNs revealed that they exhibit a behavior of implicit regularization. This ``leads to the selection [\dots] of minimum norm solutions'' (\citeauthor{Poggio.2020} \citeyear{Poggio.2020}: 30041). In other words, the SGD algorithm prefers configurations of weights that have a small norm without being forced to do so. This preference governs the algorithm's convergence and hence hypothesis choice after a rejection. The crucial point is that reaching a configuration of weights with small norm means that the values for the individual weights are small, many of them are even zero.\footnote{\textcolor{blue}{A similar behavior can be found in learning processes of other methods. For instance, SVMs are optimized for a small number of support vectors, and in Bayesian causal inference, one typically aims at finding sparse graphs. However, such behavior is commonly enforced from the outset, e.g., by a kind of explicit regularization.}} Given the structure of an ANN, this means that some nodes have a negligible or zero impact in the weighted sum feeding into a node in the subsequent layer. The flow of information is locally muted (or damped). Seen from a node receiving an input, the relevant network determining the input `looks' less complex than the actual network really is. This is why a configuration of weights with small norm is commonly interpreted as a `simple' prediction rule. Thus, a convergence to minimum norm solutions means that the algorithm is ``implicitly trying to find a solution with small `complexity' '' (\citeauthor{Neyshabur.2015} \citeyear{Neyshabur.2015}: 3).\footnote{\textcolor{blue}{While the SGD algorithm's preference for `simple' prediction rules is uncontroversial, there is some debate in ML research about whether the algorithm prefers solutions with small norm (\citeauthor{Neyshabur.2015} \citeyear{Neyshabur.2015}, \citeauthor{Poggio.2020} \citeyear{Poggio.2020}), small rank (\citeauthor{Huh.2021} \citeyear{Huh.2021}, \citeauthor{Razin.2020} \citeyear{Razin.2020}) or simple parameter-function maps (\citeauthor{VallePerez.2019} \citeyear{VallePerez.2019}). For our argumentation, the measure of simplicity to which SGD converges is less important, the point is that such a convergence takes place.}}

The behavior of the SGD algorithm contrasts sharply with any reasoning that is based on complexity measures such as the VC-dimension. We have seen above that within the standard framework of statistical learning theory, an \emph{ex ante}-simplicity of the final prediction rule is enforced by defining or optimizing for a hypothesis class that does only contain sufficiently simple prediction rules. Complexity measures can capture this type of simplicity, because their purpose is to assess the complexity of function classes. Yet in the case of ANNs, an \emph{ex post}-simplicity of the final prediction rule is enforced during the learning process in virtue of implicit regularization. Complexity measures cannot capture this type of simplicity, because it does not depend on the complexity of functions in the hypothesis class. 

Overall, the behavior of the SGD algorithm establishes the fourth parallel. In the learning process of an ANN as well as in a process of falsification, simplicity guides hypothesis choice; given their empirical adequacy, simpler hypotheses should be preferred.

Our discussion reveals that there are several profound parallels between the learning process of an ANN and a process of falsification. The discussion also reveals that despite these parallels, there are also important distinctions to be made. They clarify key components of methodological falsificationism from an ANN perspective. Yet there are further aspects that lie beyond the scope of our discussion. For instance, according to \citeauthor{Popper.1959} (\citeyear{Popper.1959}: 280), scientists should test hypotheses that make bold claims. One might question how this relates to hypotheses and hence to configurations of weights in the context of ANNs, since it is not clear how to assess their `boldness'. We consider our present contribution as a starting point for further investigation of this and other aspects that we could not address here.

\section{Conclusion}
\label{sec:conclusion}

This paper tried to shed new light on the methodology driving the learning process of ML algorithms. Whereas the field usually emphasizes the statistical side of ML and its close connection with induction, we claim that the idea of falsification is central to ML.

To argue for this claim, we closely analyzed the learning process of artificial neural networks (ANNs). The analysis revealed that statistical and algorithmic aspects are closely intertwined within the process. We discussed four parallels between the learning process of ANNs and a process of methodological falsification. We showed that the role of background knowledge is crucial in both cases. Additionally, the framework of empirical risk minimization and the behavior of implicit regularization closely parallel the two essential steps in a process of falsification: rejection of an inadequate hypothesis and replacement by a new one which is both more adequate and simpler. Consequently, the learning process of ANNs closely follows the ideas of methodological falsificationism. We showed that it even renders the latter more precise, both in quantifying what constitutes a reproducible effect and in spelling out the notions of empirical adequacy as well as simplicity. Taken together, this establishes the falsificationist account of ANNs.

The falsificationist account highlights the relevance of falsification for the methodology of ML. Contrary to previous accounts, it does not rely on complexity measures such as the VC-dimension and focuses explicitly on the `learning' component of ML. Furthermore, the account is applicable to cases in which traditional complexity measures are not meaningful. Our conjecture for future work is that the new falsificationist perspective on ANNs could be extended to other ML algorithms (e.g., SVMs) and might bring new insights into the methodology of ML, its status as a science, and the understanding of its astonishing success.

\clearpage



\end{document}